\pdfoutput=1

\documentclass[11pt]{article}

\usepackage[final]{ACL2025}

\usepackage{times}
\usepackage{latexsym}

\usepackage[T1]{fontenc}

\usepackage[utf8]{inputenc}

\usepackage{microtype}

\usepackage{inconsolata}

\usepackage{graphicx}
\usepackage{amsmath}
\usepackage{array}
\usepackage[skins]{tcolorbox}
\usepackage{multirow}
\usepackage{arydshln}
\usepackage{csquotes}
\usepackage{dashrule}

\usepackage{tabularray}

\usepackage{caption}
\usepackage{subcaption}

\definecolor{myblue}{RGB}{162, 56, 255}
\definecolor{mycyan}{RGB}{75, 186, 186}

\usepackage{float}
\usepackage{dblfloatfix}

\usepackage[skins]{tcolorbox}

\newtcolorbox{instructionframe}[2][]{%
  enhanced,colback=white,colframe=myblue,coltitle=white,boxrule=1.0pt,
  fonttitle=\mdseries,
  attach boxed title to top left={yshift=-0.5\baselineskip-0.4pt,xshift=2mm},
  boxed title style={tile,size=minimal,left=1.5mm,right=1.5mm,
    colback=myblue,before upper=\strut},
  title=#2,#1
}

\title{Synthetic Lyrics Detection Across Languages and Genres}

\author{Yanis Labrak~$^{1,2}$ \hspace{3mm} Markus Frohmann~$^{1,3}$  \hspace{3mm} Gabriel Meseguer-Brocal~$^1$ \hspace{3mm} Elena V. Epure~$^1$ \\
Deezer Research, Paris, France~$^1$ \\ LIA - Avignon University, Avignon, France~$^2$ \hspace{3mm} Johannes Kepler University Linz, Austria~$^3$ \\
\texttt{research@deezer.com}
}

\begin{document}
\maketitle
\begin{abstract}
In recent years, the use of large language models (LLMs) to generate music content, particularly lyrics, has gained in popularity. 
These advances provide valuable tools for artists and enhance their creative processes, but they also raise concerns about copyright violations, consumer satisfaction, and content spamming. 
Previous research has explored content detection in various domains. 
However, no work has focused on the text modality, lyrics, in music. 
To address this gap, we curated a diverse dataset of real and synthetic lyrics from multiple languages, music genres, and artists. 
The generation pipeline was validated using both humans and automated methods.
We performed a thorough evaluation of existing synthetic text detection approaches on lyrics, a previously unexplored data type. We also investigated methods to adapt the best-performing features to lyrics through unsupervised domain adaptation.
Following both music and industrial constraints, we examined how well these approaches generalize across languages, scale with data availability, handle multilingual language content, and perform on novel genres in few-shot settings.
Our findings show promising results that could inform policy decisions around AI-generated music and enhance transparency for users.



\end{abstract}

\section{Introduction}
\label{sec:introduction}
Recent advancements in user-friendly tools, such as Suno AI\footnote{\href{https://suno.com/}{suno.com}}, have significantly impacted the music field by introducing prompt-based interfaces that simplify music generation.
In parallel, multiple research works have been exploring
audio generation~\cite{agostinelli2023musiclmgeneratingmusictext, dhariwal2020jukeboxgenerativemodelmusic, 10536191} or lyrics generation~\cite{qian-etal-2023-unilg, nikolov-etal-2020-rapformer, tian-etal-2023-unsupervised} with impressive results.
LLMs such as GPT-4 \cite{openai2024gpt4}, Mistral 7B \cite{jiang2023mistral}, Gemma \cite{gemmateam2024gemmaopenmodelsbased}, or PaLM \cite{chowdhery2022palmscalinglanguagemodeling} have demonstrated the ability to generate human-like text without adaptation, being able to assist artists in tasks such as poem writing \cite{popescu-belis-etal-2023-gpoet} and song lyrics creation \cite{qian-etal-2023-unilg}. 

Nevertheless, the widespread use of LLMs for generating artistic content has raised concerns regarding authorship infringement \cite{novelli2024generativeaieulaw, 10.1145/3630106.3658898}, consumer satisfaction \cite{spotifyReleaseRadar}, and content spamming.
These concerns outline the need to effectively detect synthetic content to regulate its distribution and prevent misuse.
Although many methods for synthetic text detection have been proposed and explored \cite{abburi-etal-2023-simple, chen-etal-2023-token, wu-etal-2023-llmdet, pu-etal-2023-zero, wang-etal-2024-m4gt, dugan-etal-2024-raid, li-etal-2024-mage}, their effectiveness in detecting AI-generated lyrics as a form of creative content remains unclear.
Lyrics differ significantly from other text types due to their unique semantics, rhythmic structures, and socio-cultural references~\cite{E_QDC_035_0281}.
Also, existing detection benchmarks predominantly focus on English, limiting their applicability across languages, and the synthetic text used in these evaluations is often not rigorously validated.
To overcome these limitations, we propose the following contributions:
\begin{itemize}

    \item We carefully design a generation and post-processing pipeline to produce realistic lyrics, which we then validate through a human study and with automatic methods.

    \item We create and release a dataset of synthetic lyrics by using multiple generative models, featuring a wide range of lyrics for 9 languages and 18 unique music genres inspired by 1,771 artists from various countries. 

    \item We conduct extensive experiments to benchmark existing text detection approaches on this new type of synthetic text (creative and multilingual) with minimal adaptation. Our focus includes a variety of features: metrics derived from per-token probabilities in lyrics and stylistic and sentence embeddings. Then, we assess LLM2Vec \citep{behnamghader2024llm2veclargelanguagemodels} for the first time in the context of text detection, both with and without lyrics-specific adaptation, showing that it outperforms all other features on this data type.

    \item In contrast to previous works, we evaluate detectors not only for generalization to unseen generators and content (e.g., new artist style, new music genres) but also for their robustness and performance with unseen languages and varying levels of data availability in order to simulate a more realistic detection scenario.
\end{itemize}

\noindent Data, pre-processing scripts, code, and models will be publicly accessible on GitHub \footnote{\href{https://github.com/deezer/synthetic_lyrics_detection}{https://github.com/deezer/synthetic\_lyrics\_detection}} under the Apache 2.0 license and in compliance with the content copyrights.

\section{Related Work}
\label{sec:relatedwork}

The detection of machine-generated content has emerged as a well-established research domain \citep{10.5555/3053718.3053722, badaskar-etal-2008-identifying,yang2023surveydetectionllmsgeneratedcontent,9721302, 9799858, Zhou_2021_ICCV, 10.1145/3652027, 10334046}. 
Traditionally, efforts have focused on identifying generated text in areas like news \citep{bhat-parthasarathy-2020-effectively,schuster-etal-2020-limitations}, scientific writing \citep{chen-etal-2021-scixgen-scientific}, or voice spoofing in audio \citep{Wu2017ASVspoof, 9417604}. 
However, recent advances in generative models in terms of quality and creativity have underscored the need for detectors capable of identifying more complex forms of machine-generated text, such as creative content. 
In music, multiple modalities are vulnerable to AI-generated content, but current efforts have mainly targeted audio detection \cite{zang2024singfake, 7858696, afchar2024detecting}.

Detection of machine-generated text is typically framed as a binary classification task distinguishing between human-written and synthetic content \citep{liu-etal-2023-coco, huang2024tokenensemble}.
One way of solving it relies on supervised learning, where classifiers are trained based on textual encoders like RoBERTa or Longformer \cite{abdelnabi21oakland, chakraborty-etal-2023-counter, pmlr-v202-kirchenbauer23a,liu-etal-2023-coco,wang-etal-2024-m4gt,li-etal-2024-mage} or LLMs \cite{macko-etal-2023-multitude, antoun-etal-2024-text-source, chen-etal-2023-token, kumarage-etal-2023-reliable}.
This approach requires a sufficiently large training corpus, which is not always available, and may encounter overfitting issues on unseen data, including new authorial styles or generative models~\cite{uchendu-etal-2020-authorship, bakhtin2019real}.

Another line of research has focused on distinguishing between machine-generated and human-written texts using various metrics derived from output probabilities of generative models or stylistic features \cite{10.5555/3618408.3619446,su-etal-2023-detectllm, zhu-etal-2023-beat, sadasivan2024can,soto2024fewshot}. 
These methods have been proven effective, while sometimes shown to yield lower performance than the supervised ones depending on the generative model and data \cite{wang-etal-2024-m4gt,li-etal-2024-mage}.
Parallel research has explored watermark-based detection methods \cite{abdelnabi21oakland, chakraborty-etal-2023-counter, pmlr-v202-kirchenbauer23a}, but these approaches are limited by the requirement to access model logits, which is not feasible for models accessible only via APIs, such as GPT-4 \cite{openai2024gpt4technicalreport}.

As discussed above, previous research has explored content detection across various domains, yet no work has exclusively focused on the text modality, lyrics, in music. 
Moreover, prior benchmarks have primarily targeted English text and often lacked a rigorous validation of the synthetic text used in experiments, raising concerns about the findings' reliability and generalization.
These gaps highlight the need for a validated pipeline to generate and refine lyrics, the release of synthetic data that is realistic, musically diverse, and multilingual, and more targeted generalization experiments that explore various factors, including generative models, languages, and writing styles.

\section{Data Creation and Validation}
\label{sec:datasets}

As no prior public studies have addressed the detection of machine-generated lyrics, there is a lack of data reflecting the inherent diversity of song lyrics. 
To address this gap, we introduce and document the creation of the first lyrics dataset specifically designed for synthetic lyrics detection. 
This data encompasses a wide variety of artistic styles, music genres, and languages.
For generation, we chose to focus on textual input only, excluding lyrics generators that use multiple modalities, such as melody or audio \cite{qian-etal-2023-unilg, tian-etal-2023-unsupervised}. 
Likewise, we align with the most widely used tools among content creators, such as Suno and ChatGPT, which produce lyrics based entirely on text.

\subsection{Human-Written Lyrics Dataset}
\label{sec:humanlyrics}
Given the large diversity of the music catalog with lyrics from millions of artists across very different genres, styles, and languages, with new tracks being added almost every second \citep{Ingham2021}, creating a comprehensive dataset that covers these dimensions is necessary but challenging.

For this work, we curated a multilingual dataset of 3,704 human-written lyrics targeting nine languages: English (EN), German (DE), Turkish (TR), French (FR), Portuguese (PT), Spanish (ES), Italian (IT), Arabic (AR), and Japanese (JA).
The inclusion criterion was based on popularity, specifically from tracks listed in the most popular editorial playlists on an international music streaming platform\footnote{\href{https://deezer.com}{deezer.com}} as of June 2024.
Also, we ensured that each track was released within the past year and a half to minimize the possibility that the models used in the detectors had prior exposure to this content.
We evenly selected lyrics only from top-trending music genres per language, as determined by daily streaming statistics at extraction time. 
Appendix \ref{app:dataset-distribution} shows the data distribution, and Appendix \ref{appendix:genres-list} the list of popular genres per language.

To allow a quality assessment of the generated lyrics by English-speaking humans from our organization, we decided to evenly and randomly pick a sub-sample from this dataset focused on the five most popular artists from the 2023 Billboard \enquote{Top Artists}\footnote{\href{https://www.billboard.com/charts/year-end/top-artists/}{billboard.com/charts/year-end/top-artists}}, namely: Drake, Ed Sheeran, Post Malone, Taylor Swift, and The Weeknd.
Though limited in scope, this dataset is a test bed of 625 human-written lyrics (for the distribution, see Appendix \ref{app:dataset-distribution}) well-suited for assessing artistic style cloning capabilities of our LLM generation pipeline. We also use this controlled subset to identify the best detection features before running extensive experiments on robustness, scalability, and generalization.

\subsection{Synthetic Lyrics Dataset}
\label{sec:lyrics-generation}
High-quality generated text increases the difficulty of the task, providing a better evaluation and insights into a system's ability to generalize to unseen data. 
To produce human-like lyrics, we designed a four-step process that was refined through multiple iterations, with each step's output being empirically evaluated for potential issues or generation artifacts and improvements made accordingly. 
The entire pipeline is validated through a human study (Section \ref{sec:humaneval}) and an automatic evaluation focused on the regurgitation of the models (Section \ref{sec:lyrics-difference-from-real}).

\paragraph{Step 1 - Generation.}
We opted for a constrained generation with a carefully designed prompt that was short and general, including some basic formatting instructions and three lyrics examples.
The few-shot examples changed at each generation to diversify the output \citep{lu-etal-2022-fantastically} but were conditioned on the same artist for the Billboard top artists data or the same language/genre pair for the multilingual data. 
To ensure the generated lyrics closely resembled real ones, the model was instructed to follow the same  formatting guidelines as the real lyrics\footnote{\href{https://docs.lyricfind.com/LyricFind_LyricFormattingGuidelines.pdf}{docs.lyricfind.com}}.
Appendix \ref{appendix:prompt-template} shows the prompt template and Appendix \ref{sec:Hyperparameters} the hyperparameters used.

We selected four LLMs to generate varied content, ensuring their release preceded the period of the human-written lyrics. LLaMa 2 13B \cite{touvron2023llama} and 
Mistral 7B \cite{jiang2023mistral} were chosen as the foundation models. 
In particular, lyrics generated with LLaMa 2 13B were used only as training data for the Billboard top artist subset to validate generalization capabilities to new models. 
TinyLLaMa 1.1B \cite{zhang2024tinyllama} was used as a smaller, more compact model with similar performance to its corresponding foundation model. 
Lastly, we included WizardLM2 7B \cite{xu2024wizardlm}, an instruction-tuned model derived from Mistral 7B and fine-tuned on a large dataset using DPO \cite{rafailov2023direct}.

\paragraph{Step 2 - Normalization.} 
We normalized generated lyrics using regular expressions developed iteratively with each model's inclusion to remove artifacts not found in real lyrics, such as punctuation at the end of verses, quotations, references to the generation process (e.g., \enquote{here's an example of a song}), and indications of offensive content.

\paragraph{Step 3 - Initial Filtering.} 
We sampled normalized generated lyrics to match the typical style of artists or language/genre pairs using statistical metrics from real lyrics, such as sentence length, number of verses, verse size, and word count. 
Only lyrics that fell within the interquartile range of these metrics, represented by box plots created from the human-written lyrics per artist, were retained.

\paragraph{Step 4 - Semantic Similarity Filtering.} 
We performed a semantic similarity comparison between generated and human-written lyrics, retaining up to 150 synthetic lyrics that were most similar for each generative model and artist or language-genre pair. 
For this, we used the Sentence Transformers's \cite{reimers-gurevych-2019-sentence} model \texttt{all-MiniLM-L6-v2} from \citet{wang-etal-2021-minilmv2}.

\subsection{Human Evaluation}
\label{sec:humaneval}
The human evaluation aimed to assess how realistic the lyrics produced by our generation and post-processing pipeline were, providing insights into their validity.
We recruited four English-speaking subjects from our organization to determine whether 70 English lyrics from the Billboard top artists data were 'human-written' or 'machine-generated', based on text only. 
The samples were evenly split between the two classes and uniformly distributed across various artists and generative models, while subjects were unaware of this distribution to prevent bias.
Subjects also rated their confidence in each annotation on a scale from 1 to 4 (details in Appendix \ref{fig:confidence-score}). 
Post-annotation, an unstructured interview was conducted to gather insights into the decision-making process (e.g., cues used in judgments), familiarity with the lyrics, and perceived difficulty (transcribed in Appendix \ref{appendix:transcriptions}).

Table \ref{tab:human-performance} shows that the differences among subjects are substantial, with a gap of 36.9 points between the highest (ID 4) and lowest (ID 2) scores. 
The recall for the synthetic lyrics is close to or even worse than a random baseline for all the subjects except the fourth.
The detection of human-written lyrics appears better, but this might be related to a tendency to overuse this label in annotation.

\begin{table}[H]
\centering
\resizebox{\columnwidth}{!}{%
\setlength\extrarowheight{1.7pt}
\begin{tabular}{lccc}
\hline
\textit{Subject ID} & \textit{Synthetic} & \textit{Human-written} & \textit{Overall}\\ \hline
\textit{1} & 54.3 & 97.1 & 75.7 \\
\textit{2} & 40.0 & 43.4 & 41.7 \\
\textit{3} & 57.1 & 78.5 & 67.8 \\ 
\textit{4} & 74.3 & 82.9 & 78.6 \\
\hline
\end{tabular}%
}
\caption{Human subjects' recall on a sample of 70 lyrics taken from the Billboard top artists data.}
\label{tab:human-performance}
\end{table}

In Appendix \ref{app:human-eval}, we show that subjects tended to assign slightly lower confidence scores to their incorrect annotations, likely because they anticipated their mistakes to some extent.
Based on subjects' feedback detailed in Appendix \ref{appendix:transcriptions}, only one popular song by Taylor Swift was recognized.
We provide a supplementary analysis of pair inter-rater agreements in Appendix \ref{app:human-eval}.
Overall, the results highlight the task's difficulty and that the generated lyrics resemble real ones, thus validating our pipeline.

\subsection{Measuring Few-Shot Regurgitation}
\label{sec:lyrics-difference-from-real}
To ensure that the generative models used for creating our dataset do not merely reproduce the provided few-shot examples, we conducted an additional evaluation of the generated lyrics apart from the human one. 
We indexed all the human-written lyrics used to condition the models in generation with the BM25 representation \cite{10.1145/2682862.2682863}.
Then, we queried this corpus by using synthetic lyrics and checked if the few-shot examples provided as seeds in the corresponding prompt during generation scored high in this retrieval task.
Table \ref{appendix:bm25} shows that hit rates are relatively low for each rank range, indicating a low likelihood of the generated lyrics being based on the set of lyrics provided as input to condition their generation. 

\begin{table}[!htb]
\small
\centering
\setlength\extrarowheight{1.7pt}
\setlength{\tabcolsep}{10pt}
\begin{tabular}{lcc}
\hline
Rank      & \% Hit rate          & Cumulated \%  Hit rate \\
\hline
1         &  2.28                 & 2.28                   \\
2         &  1.05                 & 3.34                  \\
3         & 0.83 &  4.17                 \\
3 to 5     &  1.37                 &   5.55                \\
5 to 10    &   2.57                &  8.12                 \\
10 to 20   &  3.94                 &  12.06                 \\
20 to 50   &   7.79                &  19.86           \\     
\hline
\end{tabular}%
\caption{Hit rate (\%) by rank range when retrieving the human lyrics used as 3-shot examples during generation with the corresponding synthetic lyrics.}
\label{appendix:bm25}
\end{table}

\begin{table*}[!t]{
\small
\centering
\setlength\extrarowheight{1.0pt} 
\setlength\tabcolsep{11.4pt}
\centering
\begin{tabular}{lccccccccc}

\hline

& \multicolumn{6}{c}{\textit{Lyrics Generators}} & & & \\

\cline{2-7}

& \multicolumn{2}{c}{\text{Mistral 7B}} & \multicolumn{2}{c}{\text{TinyLLaMa}} & \multicolumn{2}{c}{\text{WizardLM2}} & \multicolumn{2}{c}{\text{Human-written}} & \\

\multicolumn{1}{r}{} &   $S$ & $U$ & $S$ & $U$ & $S$ & $U$ & $S$ & $U$ & \text{Avg.} \\
\hline

\textit{Random}  & 51.3 & 49.0 & 50.2 & 48.7 & 46.9 & 53.3 & 48.0 & 41.3 & 47.3 \\

\vspace{-3mm}\\
\cdashline{1-10}

 \vspace{-2mm}\\
 \multicolumn{10}{c}{\textit{Metrics based on LLaMa 2 7B Per-Tokens Probabilities}} \\
 \vspace{-2mm}\\

 \textit{Perplexity}  & 79.0 & 84.0 & 58.0 & 45.3 & 71.9 & 72.7 & 57.2 & 53.6 & 61.9 \\

\textit{Max.Neg Log.Lkl.}  &  75.8 & 74.3 & \textbf{77.6} & \textbf{72.3} & 63.2 & 55.7 & 83.4 & 89.4 & 78.1 \\

\textit{Shannon Entropy}  &  &  &  &  &   &  &  &  &  \\
\hspace{3mm} \textit{Max}  & 88.2 & 94.0 & 50.6 & 58.9 &  71.6 & 73.0 & 77.4 & 71.2 & 73.5 \\

\hspace{3mm} \textit{Max+Min} & 88.4 & 88.7 & 64.6 & \underline{60.2} & 68.6 & 65.3 &  80.6 & 82.8 & 77.2 \\

\textit{Min-K\%Prob (K=10)}  & \underline{92.4} & 93.7 & \underline{70.5} & 51.0 & \textbf{93.2} & \textbf{96.7} & 70.7 & 88.6 & \underline{81.3}\\

\vspace{-3mm}\\

\cdashline{1-10}

 \vspace{-2mm}\\
\multicolumn{10}{c}{\textit{Semantic and Syntactic Embeddings}} \\

\textit{SBERT}  &  &  &  &  &   &  &  &  &  \\

\hspace{3mm} \textit{MiniLMv2}  & 86.9 & 94.3 & 54.7 & 55.2 & 87.9 & 91.7 & 74.8 & 73.5 & 76.3 \\

\hspace{3mm} \textit{MPNet}   & 86.4 & \underline{95.7} & 52.0 & 51.2 & \underline{88.5} & \underline{92.7} & 82.3 & 79.7 & 79.4 \\

\textit{LLM2vec}  &  &  &  &  &   &  &  &  &  \\

\hspace{3mm} \textit{LLaMa3 8B}   & \textbf{95.1} & \textbf{96.7} & 70.0 & 59.4 & 78.3 & 80.0 &  94.7 & \underline{95.6} & \textbf{87.5} \\

\hspace{3mm} \textit{LLaMa2 7B}  & 77.8 & 88.0 & 57.5 & 45.3 &  45.1 & 48.3 & \textbf{97.6} & 90.8 & 77.3 \\

\vspace{-3mm}\\

\cdashline{1-10}

 \vspace{-2mm}\\
 \multicolumn{10}{c}{\textit{Stylistic Embeddings}} \\

\textit{UAR}  &  &  &  &  &   &  &  &  &  \\

\hspace{3mm} \textit{CRUD}   & 74.7 & 81.0 & 32.8 & 32.9 & 44.8 & 44.7 & 90.6 & 89.1 & 70.8 \\

\hspace{3mm} \textit{MUD}  & 84.2 & 88.0 & 32.7 & 37.4 & 53.2 & 59.0 &  \underline{95.4} & \textbf{95.7} & 77.3 \\

\vspace{-3mm}\\

\hline

\end{tabular}%
}
\caption{Recall scores on the Billboard top artists dataset based on various features. $S$ refers to artists seen in the vector space, and $U$ to the unseen ones. Avg. is the overall micro recall between human-written and synthetic classes. For each feature category, the best-performing one is in \textbf{bold}, and the second-best is \underline{underlined}.}
\label{tab:overall-results}
\end{table*}

\section{Lyrics Detection Experiments}
\label{sec:detectionexperiments}

We approached the detection task as a few-shot prediction using a k-nearest neighbors (k-NN) algorithm on a pre-computed lyrics features space. 
This method, which works with a limited set of lyrics, supports continuous updates as new synthetic content, including human-flagged material, becomes available. The vector space is constructed using both human-written and machine-generated lyrics, corresponding to our binary classification setup, incorporating multiple features commonly used in text detection (as detailed in Section \ref{sec:detectors-features}).
During evaluation, we applied a distance-based metric (Minkowski) to find the $k$ closest points to the input and assign the most frequent label (with $k=3$ in our experiments). 
This approach also allowed for better control and explainability by understanding the influence of individual features\footnote{While k-NN is susceptible to feature scaling, this does not pose a problem since we have full control over the features.}.

\subsection{Data Split and Evaluation Scenarios}
\label{sec:datasplit}

\paragraph{Billboard Top Artists Detection.} We extended the 625 human-written lyrics of the Billboard top artists data with 4,572 synthetic lyrics inspired by the same artists. 
To evaluate cross-artist and cross-model generalization, we reserved the lyrics from two out of five artists (The Weeknd and Taylor Swift) exclusively to assess the detector's ability to generalize to unseen authorial styles. 
The lyrics from the other artists were used for both training and evaluation splits. 
For training, we sampled 300 lyrics, evenly split between human-written and machine-generated (50 lyrics from each artist).

\vspace{1mm}

\noindent \textit{Cross-Artist and Cross-Model Generalization.} We aimed to first assess the generalization capabilities to unseen generative models (Mistral 7B, TinyLLaMa, and WizardLM2) and new artists (Taylor Swift and The Weeknd, as previously detailed).

\paragraph{Multilingual Lyrics Detection.} The dataset consists of 7,262 lyrics, with 3,558 being synthetic and 3,704 human-written, distributed across 1,771 unique artist styles. 
For training, we randomly sampled up to 5 lyrics for each class (human-written and synthetic) and each language/genre pair. 
The remaining lyrics were reserved for evaluation. 
The distribution across splits is shown in Appendix \ref{app:dataset-distribution}.
We now further discuss the evaluation scenarios.

\vspace{1mm}
\noindent 
\textit{Baseline.}
The baseline used all languages, genres, and training data to build the vector space. 

\vspace{1mm}

\noindent 
\textit{Scalability.} We varied the amount of data used to construct the vector space for the detectors, scaling the number of available lyrics from 1 to 5 per language/genre pair (108 to 540 lyrics in the vector space) and measuring the impact.

\vspace{1mm}

\noindent \textit{Cross-lingual Generalization.} We isolated a language at a time when building the vector space to evaluate how well the detector generalized when trained on a specific language and then tested on unseen languages. In particular, we assessed the detector's ability to handle unfamiliar lyrics characteristics and language-specific music genres.

\vspace{1mm}

\noindent \textit{Robustness.} We combined languages in the vector space, starting with English and gradually incorporating all 9 languages. 
This evaluated how well the detector handled multilingual data and maintained performance across diverse language inputs. 
The language order was defined by their linguistic characteristics (agglutinative, inflected, etc.) and language families (Germanic, Latin, Semitic, etc.).

\subsection{Detection Features}
\label{sec:detectors-features}
To build the vector space of human-written and synthetic lyrics, we focused on a variety of features commonly found in the literature. 

\paragraph{Probabilistic Features:}

The first group of features includes metrics derived from output probabilities of generative models. 
We took into account the segmentation of the lyrics and computed most of the metrics at the verse level, which has been experimentally proven to be more effective.
We assumed a black-box generative model to produce synthetic lyrics and relied on other models to estimate the per-tokens probabilities of the text. 
In practice, we computed those per-tokens probabilities using LLaMa 2 7B for the Billboard top artists subset. We also tested the impact of this choice by replacing LLaMa 2 7B with an alternative model, Gemma 2 9B \cite{gemmateam2024gemmaopenmodelsbased}.

\textit{Maximum Negative Log-Likelihood} \citep{10.5555/3618408.3619446, solaiman2019releasestrategiessocialimpacts, gehrmann-etal-2019-gltr, ippolito-etal-2020-automatic} calculates token-level negative log-likelihood for lyrics, treating individual verses separately. 
We took the max value across verses and use it as a 1-D feature vector for lyrics.

\textit{Perplexity}~(PPL) \cite{10.1007/978-3-319-41754-7_43} measures the overall likelihood of the lyrics based on the exponential average of the negative log-likelihood. 
In principle, lower PPL suggests the lyrics are less likely to be human-written as artistic writing could lead to higher PPL due to its unexpectedness.

\textit{Shannon entropy} \cite{Shannon1948math, 10.5555/3053718.3053722} measures the diversity or sparsity of the lyrics vocabulary based on token-level negative log-likelihood. We pooled the highest entropy value across all verses as a 1-D feature vector. 
We also considered both the highest and lowest entropy values as a 2-D feature vector to cater to the unique structure of the lyrics domain.

\textit{Min-K\% Prob} \cite{shi2024detecting} selects a sample of K\% of the lowest token-level negative log-likelihood probabilities from the entire song and averages them to create a 1-D lyrics-level feature ($K=10$ as shown in Appendix \ref{sec:min-k-impact}).

\paragraph{Semantic and Syntactic Embeddings:}

The second feature group for building the vector space includes semantic and syntactic embeddings, as differences in these aspects may exist between human-written and machine-generated lyrics \cite{jawahar-etal-2019-bert, soto2024fewshot}.
We use two models from the Sentence Transformers library (\textit{SBERT}) by \citet{reimers-gurevych-2019-sentence}: \textit{all-MiniLM-L6-v2} \cite{wang-etal-2021-minilmv2} and \textit{all-mpnet-base-v2} \cite{10.5555/3495724.3497138}. 
In addition, we also use LLM2Vec \cite{behnamghader2024llm2veclargelanguagemodels} for the first time in detection. 
LLM2Vec is an unsupervised method that transforms autoregressive LLMs into text encoders using a 3-step process: (i) enabling bidirectional attention by modifying the attention mask; (ii) masked next-token prediction (MNTP) to adapt the model to its different attention mask; and (iii, optional) SimCSE \cite{gao-etal-2021-simcse} learning to enable stronger sequence representations. The final output embedding is derived via mean-pooling. In our experiments, we used LLM2Vec models that were only tuned via MNTP since we observed that they performed the best.
In addition, we fine-tune LLM2Vec on the multilingual lyrics corpus. We refer to §\ref{sec:domain-adaptation} for details.

\paragraph{Stylistic Representations:}
The third feature group captures the authorial writing style. 
We used the Universal Authorship Representation (\textit{UAR}) model \citep{rivera-soto-etal-2021-learning} with its variants: \textit{MUD} and \textit{CRUD}, trained on data from 1 million and 5 million different Reddit users, respectively. 
\citet{soto2024fewshot} have demonstrated that these features are highly effective in distinguishing between human-written and synthetic content.

\section{Results}
In the following, we report macro-recall as the primary metric, following \citet{nakov-etal-2013-semeval,li-etal-2024-mage}. 
This ensures a realistic evaluation of the detectors, particularly since black-box models such as human predictors cannot be evaluated using AUROC. 
The focus is on minimizing false negatives for human-written lyrics and maximizing true positives for synthetic ones to prevent mislabeling.

\begin{table*}[!t]
\centering
\tiny
\resizebox{\textwidth}{!}{%
\begin{tabular}{l|l|ccccccccc|c}
\hline
 &  & \multicolumn{9}{c}{\text{Languages}} &  \\
\cline{3-11}
Scenario & Setup & EN & DE & TR & FR & PT & ES & IT & AR & JA & Avg. \\
\hline


\textit{Baseline} & All
 & 83.3 & 84.4 & 73.9 & 85.8 & 81.1 & 82.0 & 82.1 & 81.6 & 67.1 & 80.2 \\

 \cdashline{1-12}

 \multirow{5}{*}{\textit{Scalability}}
& 1 & 77.9 & 84.1 & \textbf{75.7} & \textbf{86.4} & 80.7 & 80.2 & 78.2 & 80.6 & 66.6 & 78.9 \\
& 2 & 81.2 & \textbf{84.5} & \textbf{75.7} & \underline{85.9} & 80.1 & 81.4 & 79.7 & \underline{81.6} & \underline{69.0} & 79.9 \\
& 3 & \underline{82.5} & 84.3 & 74.7 & 85.6 & \textbf{81.2} & 81.8 & 79.7 & \textbf{81.8} & \textbf{69.1} & \underline{80.1} \\
& 4 & \textbf{83.3} & 83.8 & \underline{75.2} & 85.7 & \textbf{81.2} & \textbf{82.1} & \underline{80.3} & 81.1 & 67.5 & 80.0 \\
& 5 & \textbf{83.3} & \underline{84.4} & 73.9 & 85.8 & \underline{81.1} & \underline{82.0} & \textbf{82.1} & \underline{81.6} & 67.1 & \textbf{80.2} \\

\cdashline{1-12}

\multirow{9}{*}{\textit{Cross-Lingual}}
& EN & \textbf{83.8} & 81.6 & 74.6 & 84.7 & 80.3 & 77.7 & 77.3 & 63.2 & 62.8 & 76.2 \\
& DE & \underline{70.5} & \textbf{85.7} & 74.5 & \underline{87.5} & 81.5 & \underline{81.1} & \textbf{81.5} & \underline{81.1} & 64.8 & \textbf{78.7} \\
& TR & 56.3 & 85.1 & \textbf{76.7} & 85.6 & 81.2 & 79.9 & 76.0 & 78.6 & 63.6 & 75.9 \\
& FR & \underline{70.5} & \underline{85.6} & 71.8 & \textbf{88.6} & \underline{82.3} & 80.9 & \underline{80.7} & 77.3 & 64.1 & \underline{78.0} \\
& PT & 64.4 & 69.6 & 63.2 & 70.3 & 81.8 & 74.8 & 77.3 & 55.6 & 65.6 & 69.2 \\
& ES & 68.6 & 84.8 & 75.1 & 85.1 & 80.7 & \textbf{82.3} & 79.9 & 74.9 & 62.7 & 77.1 \\
& IT & 70.1 & 83.6 & 67.6 & 85.9 & \textbf{82.7} & 80.1 & 78.8 & 68.4 & 65.1 & 75.8 \\
& AR & 54.7 & 81.7 & \underline{75.7} & 76.2 & 73.4 & 76.1 & 72.6 & \textbf{82.0} & \underline{66.7} & 73.2 \\
& JA & 69.6 & 81.5 & 68.9 & 80.3 & 80.5 & 78.7 & 74.0 & 63.7 & \textbf{68.2} & 73.9 \\

\cdashline{1-12}

\multirow{9}{*}{\textit{Robustness}}
& EN & 83.8 & 81.6 & \underline{74.6} & 84.7 & 80.3 & 77.7 & 77.3 & 63.2 & 62.8 & 76.2 \\
& + DE & \underline{84.9} & 84.3 & \underline{74.6} & 86.3 & 80.6 & 80.3 & 81.1 & 80.1 & \underline{65.6} & \underline{79.8} \\
& + TR & \textbf{85.5} & 84.3 & \textbf{75.7} & \underline{86.5} & 79.9 & 80.2 & 81.0 & 80.1 & 64.1 & 79.7 \\
& + FR & 84.8 & 84.6 & 74.2 & \textbf{87.1} & 80.6 & 80.7 & 81.3 & 79.9 & 63.9 & 79.7 \\
& + PT & 83.8 & 84.2 & 72.8 & 86.4 & 80.6 & 74.7 & 79.3 & 78.8 & 64.2 & 78.3 \\
& + ES & 83.3 & \textbf{84.8} & 73.1 & 85.5 & 78.6 & 81.4 & 81.4 & 78.4 & 63.7 & 78.9 \\
& + IT & 83.6 & \textbf{84.8} & 73.0 & 85.6 & 80.0 & \underline{82.0} & 81.5 & 78.6 & 64.2 & 79.3 \\
& + AR & 83.4 & \underline{84.7} & 72.9 & 85.6 & \underline{80.7} & \textbf{82.1} & \underline{81.8} & \textbf{82.2} & 63.4 & 79.6 \\
& + JA & 83.3 & 84.4 & 73.9 & 85.8 & \textbf{81.1} & \underline{82.0} & \textbf{82.1} & \underline{81.6} & \textbf{67.1} & \textbf{80.2} \\

\hline

\end{tabular}%
}
\caption{Recall of detectors on human-written and machine-generated lyrics in each of the four scenarios. Results reported in \textbf{bold} are the best ones for the language/scenario pairs, while the second best is \underline{underlined}.}
\label{tab:all-results}
\end{table*}

\subsection{Billboard Top Artists Detection}

We observe in Table~\ref{tab:overall-results} that no single detection feature excels equally across all generators.
However, the best feature for each group appears to be Max Negative Log Likelihood, LLM2Vec embeddings with LLaMa 3 8B, and UAR-MUD embeddings.
For the multilingual experiments, we thus used only these features.
We also observe substantial differences among features in their ability to correctly label human-written lyrics. 
The features outlined earlier as the best are particularly more accurate for human-written lyrics, too. 

Despite LLM2Vec embeddings built from LLaMa 2 7B being the most accurate for human-written lyrics, it is not the overall most effective embeddings-based method.
It is worth noticing that LLaMa 3 8B outperforms LLaMa 2 7B by an overall difference of 10.2 points. 
These LLM2Vec detectors significantly surpass others, including UAR embeddings, previously considered in the literature~\cite{soto2024fewshot} as more effective compared to earlier methods like probabilistic approaches or SBERT. For UAR, MUD performs better than CRUD by 6.5 points, highlighting the benefits of using embeddings built from more diverse data.

The performance difference during the evaluation between artists seen ($S$) in the vector space and those unseen ($U$) depends on the generator and detection features used. Unsurprisingly, artists not represented in the vector space tend to perform worse overall than those who are not. 

For generators, TinyLLaMa is less frequently detected.
On the other hand, foundation models like Mistral 7B or the instruction-tuned model are more frequently detected by both probabilistic and embeddings-based methods, indicating a worse generalization than other types of models that are aimed at human-like interactions.


To identify the bias produced by using a single model for per-token probabilities, we repeated the experiments with Gemma 2 9B (c.f. Appendix \ref{sec:overall-results-gemma}). Trends were similar to LLaMa 2 7B, yet most methods showed a performance drop. Maximum negative log-likelihood declined by 9.7 points, while Min-K\% by 27.6 points.



We also replaced k-NN with a fully-supervised multi-layer perceptron for classification.
Slight performance improvements, averaging an increase of 2.02 points, were observed in 7 out of the 8 methods,
as shown in Appendix \ref{appendix:mlp-classifier}. Still, in one instance, there was a substantial performance drop of 10.8 points, making the prediction nearly random.
The minimal performance improvement does not sufficiently justify the loss of explainability associated with using a multilayer perceptron for our task.

\subsection{Multilingual Lyrics Detection}

The baseline's detection performance varies across languages, with French performing best, followed by German (-1.4), English (-2.5), and Italian (-3.7).
More detailed results of each detection feature per language are shown in Appendix \ref{app:per-feature-multilingual}.

In terms of scalability, overall performance improves with more data points per language/genre pair, though the impact is modest, with a variance of 1.3 points between the lowest and highest scores.
Performance slightly decreases with 4 lyrics per pair or in specific languages during the scalability evaluation, with Turkish and French which lost 1.8 and 0.6 points, respectively, when moving from 1 to 5 lyrics per pair. 
Conversely, languages such as English and Italian see significant improvements, with increases of 5.4 and 3.9 points, respectively.

In terms of cross-lingual generalization, building a vector space from a single language tends to generalize well to the other 8 languages. However, vector spaces based on Portuguese, Japanese, and Arabic underperform, showing recall differences of -9.5, -4.8, and -5.5 points, respectively, compared to the best-performing language, German. 
In contrast, vector spaces based on German and French generalize well to other languages, with French frequently being the second-best source language.

Regarding robustness, including more languages in the vector space incrementally improves overall performance, increasing from 76.2\% to 80.2\% with all 9 languages (+4.0). 
However, specific languages show decreased performance when added, like Portuguese (-1.4). Turkish, French, and Arabic perform better when they are lastly integrated.

\begin{table}[t]
\vspace{-4mm}
\centering
\small
\setlength{\tabcolsep}{17pt}
\begin{tabular}{clc}
\hline

Lang                & Genre         & Score \\ \hline

\multicolumn{3}{c}{\textit{Vector Space}} \\
\cdashline{1-3}

\multirow{6}{*}{EN}
&  pop  &  86.2 \\
&  hip-hop  &  83.4 \\
&  alternative  &  82.9 \\
&  rock  &  79.6 \\
&  electronic  &  84.2 \\
&  r\&b  &  86.7 \\

\hline

\multicolumn{3}{c}{\textit{Newer Languages}} \\
\cdashline{1-3}

\multirow{6}{*}{FR}
&  hip-hop  &  81.6 \\
&  pop  &  84.1 \\
&  \textbf{french}  &  \textbf{91.3} \\
&  rock  &  86.0 \\
&  alternative  &  86.8 \\
&  r\&b  &  78.4 \\

\cdashline{1-3}

\multirow{6}{*}{AR}
&  \textbf{arabic}  &  \textbf{65.6} \\
&  pop  &  64.4 \\
&  electronic  &  65.8 \\
&  alternative  &  62.0 \\
&  hip-hop  &  61.2 \\
&  rock  &  60.1 \\

\cdashline{1-3}

\multirow{6}{*}{JA}
&  pop  &  68.0 \\
&  \textbf{asian}  &  \textbf{61.6} \\
&  rock  &  61.6 \\
&  \textbf{soundtrack}  &  \textbf{54.8} \\
&  electronic  &  60.6 \\
&  alternative  &  70.4 \\

\hline

\end{tabular}%
\vspace{-1mm}
\caption{Recall when the vector space is built on EN data and tested on unseen language and genres (in \textbf{bold}).}
\label{tab:new-music-genres}
\vspace{-3mm}
\end{table}

For the genre novelty experiment (Table \ref{tab:new-music-genres}), results show no consistent trend across all languages. 
However, lyrics from the new genre in French are detected the best, while those in Arabic and Japanese less good. 
A similar trend is observed with seen genres, where English performs better as a source language for linguistically closer languages like French but not for others.
This observation aligns with previous work \citep{epure-etal-2020-modeling} showing that the perception of the same genre varies significantly across cultures.

\subsection{Towards Evaluating Domain Adaptation}
\label{sec:domain-adaptation}

\begin{figure}[t]
    \vspace{-6mm}
    \centering
    \includegraphics[width=1\linewidth]{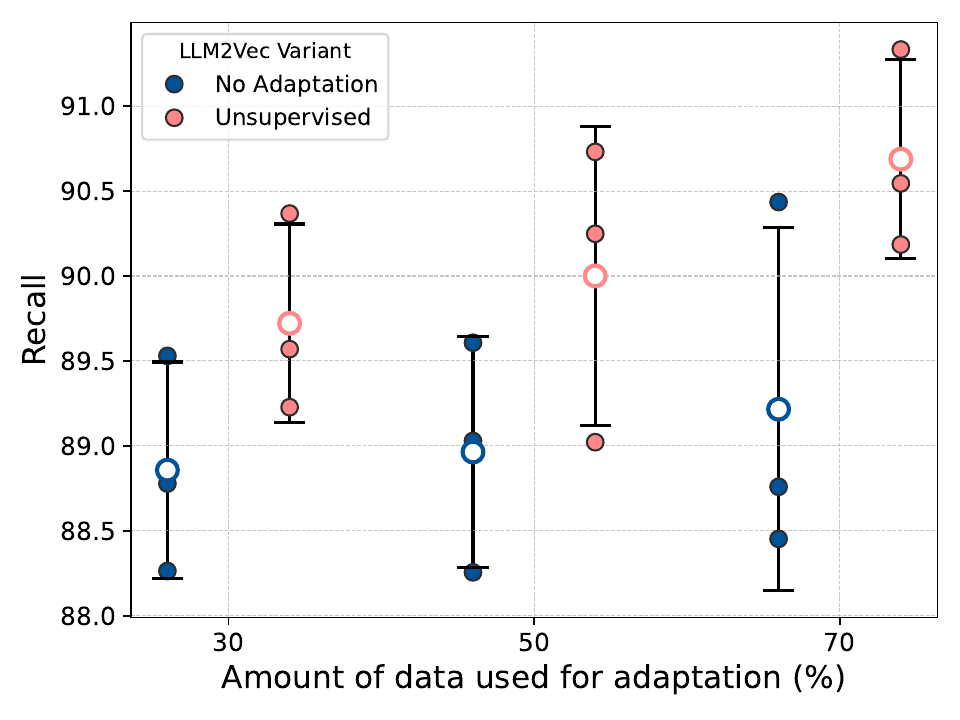}
    \vspace{-3mm}    
    \caption{Effect of domain adaptation using additional samples from the evaluation set on 3 seeds (solid circles indicate individual runs), including mean (open circle) and standard deviation. \textit{No adaptation} corresponds to the original LLM2Vec model, whereas \textit{Unsupervised} performs MNTP-based adaptation. In each scenario, we use Llama 3 8B.}
    \label{fig:ft-avg}
    \vspace{-3mm}
\end{figure}

Since the domain of lyrics highly differs from other forms of text, we now assess the effect of domain adaptation. We do so using our overall best-performing model, LLM2Vec (Llama 3 8B), in an unsupervised fashion.\footnote{We also experimented with supervised adaptation, optimized end-to-end on the task, but it consistently fell short, assumably due to insufficient generalization.} We start from the MNTP-tuned LLM2Vec model and further fine-tune it via LoRA \citep{hu2022lora} and continue tuning it via  MNTP \citep{behnamghader2024llm2veclargelanguagemodels}. To the best of our knowledge, we are the first to experiment with MNTP for unsupervised domain adaptation. The resulting domain-adapted model can be used instead of any other embeddings-based model using our existing pipeline, similarly relying on kNN-based classification. For details regarding fine-tuning experiments, we refer to Appendix \ref{appendix:domain-adaptation-hyperparameters}.

Our initial training dataset, consisting of only 525 songs from diverse genres and languages, is relatively small for domain adaptation. To address this, we expand the training dataset by incorporating additional samples, selected from the same source as the evaluation dataset but removed from the test set before inclusion. We use three different seeds for sampling. Furthermore, we evaluate the impact of corpus size on adaptation performance by varying the proportion of added samples (30\%, 50\%, 70\%, respectively). Importantly, we stratify by genre and language to ensure consistent distribution across all training and evaluation splits. For building the vector space, we rely exclusively on samples from the original training dataset, isolating the impact of domain-adaptive data on kNN-based classification and adaptation.

As shown in Figure \ref{fig:ft-avg}, MNTP-based domain adaptation appears to outperform the base LLM2Vec model with no adaptation to the lyrics domain, with the gap seeming to increase with the size of the training dataset. The difference is particularly stark in some languages, such as Japanese, as shown in Appendix \ref{appendix:domain-adaptation-lang-results}.

\section{Conclusion}


In this work, we presented a diverse dataset of lyrics to evaluate detectors' generalization capabilities. 
We then conducted a quantitative evaluation over various scenarios to assess detectors' robustness, capabilities to scale, and generalizability across languages and new genres.
The results show that our generation pipeline produces lyrics that are very difficult to distinguish by humans from real ones, thus validating it.
Using automated methods, the detection performance varies greatly depending on the LLM used for lyrics generation as well as the type of feature and artistic styles used when building the embedding space.
Increasing the amount of training data only marginally improves detection performance, whereas expanding the number of languages has a more potent impact; cross-lingual performance of detectors is highly dependent on the source language.
We adapted the best-performing features, based on LLM2Vec, to the distinct features of the lyrics domain via novel unsupervised means, indicating that MNTP-based unsupervised domain adaptation improves detection performance.
Overall, our dataset and detection experiments pave the way for more robust detection of AI-generated music, thereby enabling improved fairness in the music industry.

\section{Ethical Considerations}

Revealing the weaknesses of systems (challenging languages or music genres) can enable malicious actors to exploit these vulnerabilities further and create content that capitalizes on these flaws, such as generating and publishing machine-generated content that is harder to detect on music streaming platforms.
However, exposing these limitations to the scientific community is crucial for a better understanding of the methods and for enhancing them in future iterations.

Regarding the human study, the subjects were
recruited from our organization and performed
the annotation during their regular paid hours.
The participation in the study was on a voluntarily
basis.

\section{Limitations}

Our study has several limitations. 
Firstly, the rapid evolution of models poses a challenge, as future LLMa might generate highly diverse and unpredictable human-like lyrics, potentially outdating our detectors. Secondly, our choice of languages is limited. 
We do not know how our systems and lyrics generators will perform with sparse or under-represented languages or specific dialects. 
Additionally, we have not tested how these systems handle typos, grammatical, or semantic errors. Other factors, such as the impact of genre, tenses, or the source of the lyrics, are also still underexplored.

Moreover, we have not tested the effect of scaling data for unsupervised adaptation to millions of songs due to limited availability.

Lastly, conducting the human validation step on a larger dataset, incorporating a broader range of languages and participants from diverse socio-economic backgrounds, would provide valuable insights into the quality of the synthetic data used for generalization assessment. However, due to the limited number of subjects and the restricted language diversity within the group, we were unable to carry out this additional evaluation for now.
\bibliography{ACL2025}

\appendix

\section{Data Distribution}
\label{app:dataset-distribution}


The source of the lyrics is mentioned as either $H$ for human-written or $G$ for generated. The explicit genre names associated with denominations $G1$ to $G6$ are listed in Appendix \ref{appendix:genres-list}. The backslash character separating both figures from the same Language/Source/Genre triplet refers to the number of lyrics available in the vector space ("train") and test subsets, respectively.

\begin{table}[H]
\centering
\setlength\extrarowheight{1.7pt}
\resizebox{\columnwidth}{!}{%
\begin{tabular}{ccccccccc}
\hline
&        & \multicolumn{6}{c}{Genre}                           &          \\ \cline{3-8}

Lang                & Source & G1     & G2     & G3     & G4     & G5     & G6     & All      \\ \hline

\multirow{2}{*}{EN} & H      & 5 / 75 & 5 / 75 & 5 / 75 & 5 / 75 & 5 / 75 & 5 / 75 & 30 / 450 \\
& G      & 5 / 75 & 5 / 75 & 5 / 75 & 5 / 75 & 4 / 75 & 4 / 75 & 28 / 450 \\ \hline

\multirow{2}{*}{DE} & H      & 5 / 75 & 5 / 48 & 5 / 44 & 5 / 75 & 5 / 75 & 5 / 75 & 30 / 392 \\
& G      & 5 / 75 & 5 / 75 & 5 / 75 & 5 / 75 & 5 / 75 & 5 / 75 & 30 / 450 \\ \hline

\multirow{2}{*}{TR} & H      & 5 / 75 & 5 / 12 & 5 / 27 & 5 / 75 & 5 / 75 & 5 / 75 & 30 / 339 \\
& G      & 4 / 38 & 2 / 8  & 1 / 2  & 5 / 75 & 5 / 60 & 5 / 58 & 22 / 241 \\ \hline

\multirow{2}{*}{FR} & H      & 5 / 75 & 5 / 75 & 5 / 75 & 5 / 75 & 5 / 75 & 5 / 75 & 30 / 450 \\
& G      & 5 / 75 & 5 / 75 & 5 / 75 & 5 / 75 & 5 / 75 & 5 / 75 & 30 / 450 \\ \hline

\multirow{2}{*}{PT} & H      & 5 / 75 & 5 / 75 & 5 / 75 & 5 / 75 & 5 / 75 & 5 / 75 & 30 / 450 \\
& G      & 5 / 75 & 5 / 75 & 5 / 75 & 5 / 75 & 4 / 75 & 5 / 75 & 29 / 450 \\ \hline

\multirow{2}{*}{ES} & H      & 5 / 75 & 5 / 75 & 5 / 75 & 5 / 75 & 5 / 75 & 5 / 75 & 30 / 450 \\
& G      & 5 / 75 & 5 / 75 & 5 / 75 & 5 / 75 & 5 / 75 & 5 / 75 & 30 / 450 \\ \hline

\multirow{2}{*}{IT} & H      & 5 / 8  & 5 / 5  & 5 / 75 & 5 / 10 & 5 / 75 & 5 / 38 & 30 / 211 \\
& G      & 5 / 3  & 5 / 4  & 4 / 39 & 3 / 1  & 5 / 75 & 5 / 28 & 27 / 150 \\ \hline

\multirow{2}{*}{AR} & H      & 5 / 58 & 5 / 75 & 5 / 68 & 5 / 46 & 5 / 75 & 5 / 32 & 30 / 354 \\
& G      & 5 / 73 & 5 / 75 & 5 / 75 & 5 / 75 & 5 / 75 & 5 / 34 & 30 / 407 \\ \hline

\multirow{2}{*}{JA} & H      & 5 / 18 & 5 / 75 & 5 / 40 & 5 / 75 & 5 / 75 & 5 / 55 & 30 / 338 \\
& G      & 4 / 6  & 5 / 75 & 5 / 22 & 5 / 54 & 5 / 75 & 5 / 23 & 29 / 255 \\ \hline

&       &  &   &   &  &  & Total & 525 / 6,737 \\ \hline
\end{tabular}%
}
\caption{Distribution of the multilingual data across languages.}
\label{tab:stats-data}
\end{table}

Considering the billboard top artists subset, the distribution is as follows:

\begin{table}[H]
\centering
\small
\resizebox{\columnwidth}{!}{%
\setlength\extrarowheight{1.7pt}
\begin{tabular}{lrcc}
\hline
 & \text{Artists}      & \text{Generated}  & \text{Human-written} \\ \hline

\multicolumn{4}{c}{\textit{Vector Space ("Train")}} \\

\multirow{3}{*}{\textit{Seen ($S$)}}
& \textit{Drake}        & 50$^\dagger$                 & 50            \\
& \textit{Post Malone}  & 50$^\dagger$                 & 50            \\
& \textit{Ed Sheeran}   & 50$^\dagger$                 & 50            \\

\hline

\multicolumn{4}{c}{\textit{Evaluation ("Test")}} \\

\multirow{3}{*}{\textit{Seen ($S$)}}
& \textit{Drake}        & 931                 & 128            \\
& \textit{Post Malone}  & 769                 & 42            \\
& \textit{Ed Sheeran}   & 902                 & 84            \\

\cdashline{2-4}

\multirow{2}{*}{\textit{Unseen ($U$)}}
& \textit{Taylor Swift} & 922                 & 153            \\
& \textit{The Weeknd}   & 898                 & 68            \\ \hline

\multicolumn{2}{r}{\text{Total}}   & \text{4,572}       & \text{625}  \\ \hline
\end{tabular}
}
\caption{Distribution of the billboard top artists subset. }
\label{tab:corpus-distribution}
\end{table}

\section{Music Genres Per  Language}
\label{appendix:genres-list}

The language-specific genre acronyms refer to the following genres (each according to its language):

\begin{table}[H]
\centering
\setlength\extrarowheight{1.7pt}
\resizebox{\columnwidth}{!}{%
\begin{tabular}{ccccccc}
\hline
Lang & G1          & G2            & G3         & G4                     & G5            & G6         \\ \hline
FR   & alternative & french & hip-hop   & pop                    & r\&b          & rock       \\
IT   & alternative & electronic    & hip-hop   & jazz                   & pop           & rock       \\
ES   & alternative & electronic    & hip-hop   & latin-american & pop           & rock       \\
TR   & alternative & electronic    & folk       & hip-hop               & pop           & rock       \\
EN   & alternative & electronic    & hip-hop   & pop                    & r\&b          & rock       \\
DE   & alternative & edm           & electronic & hip-hop               & pop           & rock       \\
PT   & christian   & hip-hop      & mpb        & pop                    & samba-pagode & sertanejo  \\
JA   & alternative & asian  & electronic & pop                    & rock          & soundtrack \\
AR   & alternative & arabic & electronic & hip-hop               & pop           & rock       \\ \hline
\end{tabular}%
}
\caption{Genres selected for each of the nine languages, where "mpb" refers to \enquote{Música popular brasileira}.}
\label{tab:list-genres}
\end{table}

\section{Prompt Template}
\label{appendix:prompt-template}

Figure \ref{prompt:template} displays the prompt template used to generate lyrics with 3-shot in-context learning based on human-written lyrics:

\begin{figure}[htb]
\centering
\scriptsize
\begin{instructionframe}{3-shot Lyrics Generation Template}
\textit{Example 1:}

\vspace{1mm}

\texttt{\{\{lyrics 1\}\}}

\vspace{1mm}

\textit{Example 2:}

\vspace{1mm}

\texttt{\{\{lyrics 2\}\}}

\vspace{1mm}

\textit{Example 3:} 

\vspace{1mm}

\texttt{\{\{lyrics 3\}\}}

\vspace{1mm}

\textit{Lyrics rules:}

\vspace{1mm}

- The lyrics should be structure in optional stanzas like “Verse”, “Chorus” and “Bridge” 

\vspace{1mm}

- The beginning of each line should start with a capital letter. 

\vspace{1mm}

- Do not use repeat tags to signify if a line or stanza is repeated. Instead, write each line or stanza however many times it is said. 

\vspace{1mm}

- Do not write out any sounds that are heard in the song, like “gunshot”, “clap”, “horn”, etc. 

\vspace{1mm}

- Remove all labels such as [Talking], {Speaking}, or (Whispering). 

\vspace{1mm}

- Any word cut short should have one apostrophe in place of the missing letters. For example: givin’, livin’. 

\vspace{1mm}

- Slang is acceptable but the artist must pronounce it that way. Slang should only be used if the word sounds differently than the grammatically correct word. For example, “for shizzle” can be used but “becuz” should be spelled “because”. 

\vspace{1mm}

- Exaggerations should be cut down to the original word or punctuation. For example, “ohhhh” should be “oh” and “bang!!!!!” should be “bang!” 

\vspace{1mm}

- Background vocals should be placed on the same line they’re said but in parentheses. For example, “I’m a survivor (What, what)” 

\vspace{1mm}

- Prevent using too much background vocals 

\vspace{1mm}

\textit{Generate a new lyrics based on the style of what \enquote{\texttt{\{\{artist name\}\}}} is doing and don't mention me the fact that the lyrics is offensive:}
\end{instructionframe}
\caption{3-shot lyrics generation template.}
\label{prompt:template}
\end{figure}

\section{Lyrics Generation Hyperparameters}
\label{sec:Hyperparameters}


Table \ref{tab:llm-hyperparameters} lists all the hyperparameters used during the lyrics generation process to ensure reproducibility.
All models were quantized in GGUF Q4 to run with a reasonable inference time on consumer-grade hardware to replicate real-world usages.
We used 3 NVIDIA RTX A5000 24GB GPUs for all our experiments.

\begin{table}[H]
\small
\centering
\setlength\extrarowheight{1.7pt}
\setlength{\tabcolsep}{20pt}
\begin{tabular}{lc}
\hline
Parameter & Value \\
\hline
temperature & 0.8 \\
top\_k & 40 \\
top\_p & 0.9 \\
num\_predict & 2048 \\
quantization & Q4\_0 \\
seed & 42 \\
\hline
\end{tabular}%
\caption{Hyperparameters for the lyrics generator LLMs.}
\label{tab:llm-hyperparameters}
\end{table}

\section{Confidence Score in Human Study}
\label{fig:confidence-score}

Figure \ref{prompt:confidence-options} lists confidence score options and their descriptions provided to the subjects during the annotation task.

\begin{figure}[H]
\centering
\scriptsize
\begin{instructionframe}{Confidence scores options}

1 = Willing to defend my annotation, but it is fairly likely that I missed some details.

\vspace{2mm}

2 = Pretty sure, but there’s a chance I missed something. Although I have a good feel for this area in general, I did not carefully check the lyrics details.

\vspace{2mm}

3 = Quite sure. I tried to check the important points carefully. It’s unlikely, though conceivable, that I missed something that should affect my annotation.

\vspace{2mm}

4 = Positive that my annotation is correct. I read the lyrics very carefully.

\end{instructionframe}
\caption{List of confidence scores options and their descriptions.}
\label{prompt:confidence-options}
\end{figure}

\section{ Human Evaluation}
\label{app:human-eval}

Table \ref{tab:confidence} highlights that subjects tended to assign slightly lower confidence scores to their incorrect annotations, likely because they anticipated their mistakes to some extent. This is most noticeable in Subject 3, who exhibits a 31.5\% gap in confidence.
\begin{table}[H]
\centering
\small
\begin{tabular}{l|cccc}
\hline
\textit{Subject ID}  &  \textit{1} & \textit{2} & \textit{3} & \textit{4} \\ \hline
\textit{Incorrect} & 3.3 & 2.1 & 1.9 & 2.4 \\
\textit{Correct} & 3.4 & 2.2 & 2.5 & 2.4 \\
\hline
\end{tabular}%
\caption{Confidence scores, averaged for incorrect and correct annotations for each subject.}
\label{tab:confidence}
\end{table}

Table \ref{tab:inter-participants-agreement} shows that subjects fully agreed 28.57\% of the time, while in 71.43\% of cases, at least one disagreed. 
This led to lower Cohen's Kappa and Gwet's AC1 values, reflecting the task's difficulty and participant divergence. 
Kappa scores involving Subject 2 were near or worse than random, with negative Kappa and Gwet's AC1 values. 

\begin{table}[H]
\centering
\small
\setlength\extrarowheight{1.7pt}
\begin{tabular}{lccc}
\hline

\textit{Subject Pair}     & \multicolumn{1}{c}{$\kappa$} & \multicolumn{1}{c}{$\mathcal{G}$} & \multicolumn{1}{c}{\textit{Agreement}} \\ \hline

\textit{1  \&  2 } &  3.53  &  15.47  &  54.29 \\
\textit{1  \&  3 } &  29.81  &  43.75  &  68.57 \\
\textit{1  \&  4 } &  35.46  &  41.04  &  68.57 \\
\textit{2  \&  3 } &  17.85  &  22.28  &  60.00 \\
\textit{2  \&  4 } &  -9.29  &  -7.78  &  45.71 \\
\textit{3  \&  4 } &  30.52  &  32.80  &  65.71 \\

\hline


\end{tabular}%
\caption{Inter-participants agreement statistics. $\kappa$ is referring to Cohen's Kappa and $\mathcal{G}$ to Gwet's AC1.}
\label{tab:inter-participants-agreement}
\end{table}

\section{Transcribed Human Interviews}
\label{appendix:transcriptions}

We requested the participants to answer three questions after completing the annotation of the 70 lyrics to gather their feedback on the task they performed. All the transcribed interviews are listed in Figure \ref{appendix:participants-feedback}:

\begin{table*}
\centering
\small
\setlength\extrarowheight{2pt}
\begin{tabular}{lccccccccc}
\hline

 & \multicolumn{6}{c}{\textit{Lyrics Generators}} &  & &  \\

\cline{2-7}

 \multicolumn{1}{r}{}  & \multicolumn{2}{c}{\text{Mistral 7B}} & \multicolumn{2}{c}{\text{TinyLLaMa}} & \multicolumn{2}{c}{\text{WizardLM2}} & \multicolumn{2}{c}{\text{Human-written}} &  \\

\multicolumn{1}{r}{} & $S$ & $U$ & $S$ & $U$  & $S$ & $U$  & $S$ & $U$ & \text{Avg.} \\
\hline


\textit{Perplexity} & 46.2 & 57.0 & 50.2 & 41.1 & 47.8 & 48.3 & 57.1 & 53.7 & 52.2 \\

\textit{Max. Neg. Log-Likelihood}   &  57.3 & 53.3 & \underline{61.9} & 56.0  & 54.5 & 50.7 & 49.4 & 52.4 & 53.0 \\

 \textit{Shannon Entropy}    &   &  &   &  &   &  &   &  &  \\

\hspace{3mm} \textit{Max} &  \underline{82.4} & \textbf{88.0} & 53.1 & 57.7  & \textbf{66.0} & \textbf{73.7} & \underline{74.8} & \underline{62.3} & \underline{70.4} \\

\hspace{3mm} \textit{Min+Max} &  \textbf{84.0} & \textbf{88.0} & \textbf{64.2} & \textbf{63.9}  & \underline{61.3} & \underline{72.0} & \textbf{83.2} & \textbf{72.8} & \textbf{76.3} \\

 \textit{Min-K\% Prob (k=10)} & 47.8 & 52.0 & 51.5 & \underline{61.7} & 47.1 & 43.7  & 58.0 & 50.0 & 53.7 \\

\hline

\end{tabular}%
\caption{Recall scores on the billboard top artists subset for detectors based on probabilistic features computed using Gemma 2 9B rather than LLaMa 2 7B. $S$ refers to the artists seen in the vector space and $U$ to the unseen ones. Avg. is the overall micro recall score between human-written and machine-generated classes.}
\label{tab:overall-results-gemma}
\end{table*}

\begin{figure}[H]
\centering
\scriptsize
\begin{instructionframe}{Participant's Feedback}

\textbf{Q1: Can you write me a short explanation of what do you refer to when you were labeling the lyrics ? Which characteristics have motivated your choices ?}


\textbf{Answer P1:} I was looking to multiple characteristics, such as if the refrain is every time the same or not, the rhythms at the end of the sentences, the sparsity of the words used at the beginning of the sentences or the overall structure of the lyrics.

\vspace{2mm}

\textbf{Answer P2:} I expected lyrics to be generated if there was too much repetition, excessive punctuation (particularly too many commas within the verses), very few rhymes, or if the length of the lyrics was excessively long. 

\vspace{2mm}

\textbf{Answer P3:} Generally, I started by looking at the structure of the lyrics. Which paragraph corresponds to the choruses, whether the verses are of similar length or not, and whether there is a visible structure that stands out. If no particular structure stood out, I focused on the coherence of the lyrics. If there was a noticeable structure, I also looked at the rhymes and the progression of the story verse by verse. If the rhymes were poorly done/strange or of uneven quality, if the verses were too unbalanced, if lyrics from the verses were repeated in the choruses, or if there was not much difference between a verse and a chorus, I tended to consider it as machine-generated.

\vspace{2mm}

\textbf{Answer P4:} The main point for me is the song’s structure. Machine-generated lyrics often have a more poetic than lyrical structure. The variations of the chorus were another key indicator, in particular, machine-generated lyrics tend to create many different versions. Another hint for me was the use of counterpoints (usually in parentheses), which machine-generated lyrics tend to overuse. Finally, whenever the topic of the lyrics was explicit, it was definitely a human-written lyric, since machine are not conditioned to generate such content.

\noindent
\hdashrule[0.5ex]{7cm}{0.1pt}{2mm}

\textbf{Q2: Have you been able to recognize one or more songs during the annotation ?}


\textbf{Answer P1:} Yes, one song "Red" by Taylor Swift.

\vspace{2mm}

\textbf{Answer P2:} 1 song from Taylor Swift 

\vspace{2mm}

\textbf{Answer P3:} I had the feeling that I recognized two other songs. In those cases, I gave a rating of maximum confidence. 

\vspace{2mm}

\textbf{Answer P4:} Yes, two.

\noindent
\hdashrule[0.5ex]{7cm}{0.1pt}{2mm}

\textbf{Q3: Do you consider it as difficult task and why ? (short answer only)}


\textbf{Answer P1:} Yes, it is difficult to get confident on some lyrics since I am not used to focusing on the lyrics when listening to a song.

\vspace{2mm}

\textbf{Answer P2:} Yes, especially the rap and hip hop songs. The lyrics were very convincing and often I felt like guessing the answer with no real idea of what to choose.

\vspace{2mm}

\textbf{Answer P3:} I found this task relatively difficult (as shown by my confidence score), so yes.

\vspace{2mm}

\textbf{Answer P4:} Yes. Most of the topics are coherent and follow a natural story telling. Rhymes are also nice. So I needed to focus on other aspects.

\end{instructionframe}
\caption{Transcribed interview in the human study.}
\label{appendix:participants-feedback}
\end{figure}


\section{Gemma-Based Per-Token Probabilities}
\label{sec:overall-results-gemma}

To check the potential impact on the results when using another model to compute per-token probabilities, we conducted the same experiments with the Gemma 2 9B model. 
Similar patterns to those seen with LLaMa 2 7B were observed, though most features showed a performance decline as shown in Table \ref{tab:overall-results-gemma}. 
In particular, the maximum negative log-likelihood and Min-K\% probabilities methods were significantly impacted, with a 9.7 and 27.6 points drop, respectively, due to the model's reduced ability to distinguish between human-written and machine-generated content.

\section{Min-K \% Prob - Impact of K}
\label{sec:min-k-impact}

In order to understand the impact of the $K$ value on the detection performance, we decided to perform an exhaustive search over the values of $K$ as seen in Table \ref{tab:min-k-values}. In the case of our specific data, we observe an optimal $K$ value at $10$. 

\begin{table}[H]
\centering
\small
\setlength\extrarowheight{1.7pt}
\begin{tabular}{cc}
\hline
Min-K\% (\%) & Recall     \\
\hline
5            & 77.0    \\
10           & 79.2    \\
20           & 73.5      \\
30           & 64.3      \\
40           & 59.0     \\
50           & 57.0      \\
60           & 53.4      \\
70           & 52.7      \\
80           & 52.9   \\
\hline
\end{tabular}%
\caption{Overall recall on the test set for the Min-K\% Prob detector according to the selected K value.}
\label{tab:min-k-values}
\end{table}

\section{Results for the Multi-layer Perceptron Classifier}
\label{appendix:mlp-classifier}

An average performance gain of 2.02 points was seen in 7 of the 8 methods (limited sub-sample of methods) when replacing k-NN with a multilayer perceptron, as shown in Table \ref{tab:mlp-results}. However, the perplexity-based method experienced a 10.8 points drop, making predictions almost random.

\begin{table}[h]
\small
\centering
\begin{tabular}{lccc}
\hline
Method                   & k-NN  & MLP   & Diff. \\ \hline

Max. Neg. Log-Likelihood & 82.4 & 84.1 & +1.7      \\

\textit{Shannon Entropy}    &   &  &  \\
\hspace{3mm} Max      & 75.4 & 77.1 & +1.7      \\
\hspace{3mm} Min+Max  & 80.1 & 81.9 & +1.8  \\

Perplexity           & 60.8 & 50.0 & -10.8 \\
Min-K\% Prob (k=10)      & 79.2 & 80.5 & +1.3      \\

\textit{LUAR}    &   &  &  \\
\hspace{3mm} CRUD    & 74.8 & 77.0 & +2.2      \\
\hspace{3mm}  MUD     & 79.2 & 81.7 & +2.5      \\

SBERT MiniLMv2                    & 76.1 & 79.1 & +3.0      \\ \hline

\end{tabular}%
\caption{Same experimental setup as Table \ref{tab:overall-results} except that we used a multi-layer perceptron rather than a k-NN algorithm. The reported results show the overall scores (last column of the Table \ref{tab:overall-results}).}
\label{tab:mlp-results}
\end{table}

\section{Featured-based Detection Results on the Multilingual Lyrics}
\label{app:per-feature-multilingual}








\begin{table}[h]
\small
\resizebox{\columnwidth}{!}{%
\begin{tabular}{cccc}

\hline

& \multicolumn{3}{c}{Methods} \\

\cline{2-4}

\multicolumn{1}{l}{Langs} & \multicolumn{1}{l}{LLM2Vec} & \multicolumn{1}{l}{Max Neg Log Like.} & \multicolumn{1}{l}{UAR} \\

\hline

EN                        & 90.6                        & 59.3                                  & 100.0                    \\
DE                        & 97.4                        & 56.7                                  & 99.2                    \\
TR                        & 82.7                        & 56.5                                  & 82.4                    \\
FR                        & 97.7                        & 62.1                                  & 97.6                    \\
PT                        & 89.2                        & 54.8                                  & 99.3                    \\
ES                        & 92.3                        & 54.7                                  & 99.0                    \\
IT                        & 83.0                        & 63.3                                  & 100.0                    \\
AR                        & 92.1                        & 58.9                                  & 93.6                    \\
JA                        & 71.5                        & 55.3                                  & 74.6                   \\

\hline
Avg.                      & 88.5                        & 58.0                                  & 94.0        \\
\hline

\end{tabular}
}
\caption{Per feature performances over all languages for the baseline scenario, for the best-performing detection methods. The maximum negative log likelihood is computed using LLaMa 3 8B \cite{llama3modelcard}.}
\label{tab:all-results-top-methods}
\end{table}

Table \ref{tab:all-results-top-methods} presents performances of the baseline scenario for the best-performing features in each category, namely LLM2Vec LLaMa 3 8B, Maximum Negative Log Likelihood, and UAR MUD. We can observe that they exhibit significantly different behavior across languages. Both LLM2Vec and LUAR experience minimal performance degradation across most languages except for Arabic, Turkish, and Japanese. 
Conversely, the Maximum Negative Log Likelihood features consistently underperform compared to the other two features.

\section{Experiment Details for Domain Adaptation}
\label{appendix:domain-adaptation-hyperparameters}
For unsupervised adaptation of LLM2Vec, we employ LoRA-based fine-tuning and employ the same LoRA config as \citet{behnamghader2024llm2veclargelanguagemodels} using a rank of 16, alpha of 16. and LoRA dropout of 0.05.
We use a learning rate of 5e-5, a batch size of 32, and a maximum of 512 tokens and train for 500 steps, masking out 20 \% of tokens. 

\section{Effect of $k$ in kNN}
\label{appendix:knn-k-ablation}
We have chosen the best K experimentally on a smaller validation set from the Billboard, English only data. In Table \ref{tab:knn-ablation}, we show results on the multilingual test corpus when using the LLM2Vec embeddings (we could notice a similar trend for the other detection features). 
Similar to the behaviour on the English-only dataset, increasing the K higher than 3 does not increase the scores much.

\begin{table}[H]
\small
\centering
\begin{tabular}{l|ccccc}
\hline
\textbf{Langs} & \textbf{k=1} & \textbf{k=3} & \textbf{k=5} & \textbf{k=10} & \textbf{k=20} \\ \hline
EN & 90.97 & 89.55 & 89.75 & 89.41 & 72.49 \\
DE & 97.46 & 97.61 & 98.07 & 98.14 & 98.17 \\
TR & 82.54 & 82.76 & 82.76 & 82.76 & 82.54 \\
FR & 96.84 & 97.71 & 98.14 & 98.14 & 98.15 \\
PT & 89.28 & 89.46 & 89.22 & 90.76 & 90.72 \\
ES & 94.11 & 92.33 & 92.22 & 92.00 & 92.11 \\
IT & 80.53 & 83.09 & 82.79 & 82.58 & 80.91 \\
AR & 92.01 & 92.03 & 92.62 & 92.81 & 91.43 \\
JA & 70.43 & 70.85 & 71.23 & 69.34 & 70.47 \\
Avg. & 88.24 & 88.38 & 88.53 & 88.44 & 86.33 \\ \hline
\end{tabular}
\caption{Results on the multilingual dataset with LLM2Vec + Llama3 8B when varying $k$ in kNN.}
\label{tab:knn-ablation}
\end{table}

\section{Results with AUROC}
\label{appendix:auroc_results}

\begin{table}[H]
\small
\centering
\label{tab:multilingual-results}
\begin{tabular}{l|cccc}
\hline
\textbf{Language} & \textbf{LLM2Vec} & \textbf{LUAR} & \textbf{Entropy} & \textbf{PPL} \\ \hline
EN & 96.5 & \textbf{100.0} & 99.1 & 63.3 \\
DE & 98.0 & \textbf{99.5} & 97.4 & 61.0 \\
TR & \textbf{92.9} & \textbf{92.9} & 68.4 & 58.7 \\
FR & \textbf{99.4} & 99.0 & 98.2 & 66.1 \\
PT & 97.1 & \textbf{99.6} & 99.6 & 60.2 \\
ES & 95.1 & \textbf{99.6} & 96.9 & 55.4 \\
IT & 90.4 & \textbf{100.0} & 95.2 & 60.8 \\
AR & 93.7 & \textbf{95.9} & 68.9 & 62.1 \\
JA & 80.7 & \textbf{94.1} & 87.4 & 59.4 \\ \hline
Avg. & 93.7 & \textbf{97.8} & 90.1 & 60.8 \\ \hline
\end{tabular}
\caption{Results on the multilingual dataset with AUROC using four different classifiers.}
\label{tab:auroc-results}

\end{table}

The AUROC analysis reveals distinct patterns across detection methods and languages. LUAR demonstrates superior performance (97.8\% average), particularly excelling in Indo-European languages with perfect or near-perfect scores. While LLM2Vec (93.7\% average) and the Entropy-based classifier (90.1\%) perform well on Indo-European languages, they struggle significantly with morphologically rich languages like Turkish and Arabic (around 68\% for Entropy) and different writing systems like Japanese (80.7\% for LLM2Vec). The Perplexity-based approach's consistent underperformance (60.8\% average) across all languages suggests fundamental limitations in using raw probability scores for detection.

\section{Results with Majority Voting Classifier}
\label{appendix:majority}

\begin{table}[h]
\small
\setlength\tabcolsep{5pt}
\begin{tabular}{lccccc}
\hline
 & \textbf{LLM2Vec} & \textbf{Max NLL} & \textbf{UAR} & \textbf{Entropy} & \textbf{Maj.} \\
\hline
EN & 90.6 & 59.3 & 100.0 & 96.6 & 100.0 \\
DE & 97.4 & 56.7 & 99.2 & 97.2 & 99.0 \\
TR & 82.7 & 56.5 & 82.4 & 65.0 & 82.5 \\
FR & 97.7 & 62.1 & 97.6 & 97.8 & 97.9 \\
PT & 89.2 & 54.8 & 99.3 & 99.1 & 99.4 \\
ES & 92.3 & 54.7 & 99.0 & 95.0 & 97.3 \\
IT & 83.0 & 63.3 & 100.0 & 95.9 & 98.0 \\
AR & 92.1 & 58.9 & 93.6 & 65.8 & 93.8 \\
JA & 71.5 & 55.3 & 74.6 & 86.2 & 78.7 \\
\hline
Avg. & 88.5 & 58.0 & 94.0 & 88.7 & 94.1 \\
\hline
\end{tabular}
\caption{Per feature performances over all languages for the baseline scenario with a majority voting classifier, combining votes from the 4 best-performing classifiers, which are also shown for clarity. }
\label{tab:majority-vote}
\end{table}

The majority voting approach (Maj.) achieves the highest average performance at 94.1\%, showing only marginal improvement over UAR at 94.0\%. This minimal gain suggests that combining multiple classifiers through majority voting does not provide substantial benefits over the best individual classifier (UAR). The similar performance between majority voting and UAR also suggests that the different detection methods might be capturing similar features or making correlated errors, limiting the potential benefits of ensemble approaches.

\section{Per-language Domain Adaptation  Results}
\label{appendix:domain-adaptation-lang-results}

\begin{figure*}[h]
    \vspace{-6mm}
    \center
    \begin{subfigure}{0.45\textwidth}
        \centering
        \includegraphics[width=\textwidth]{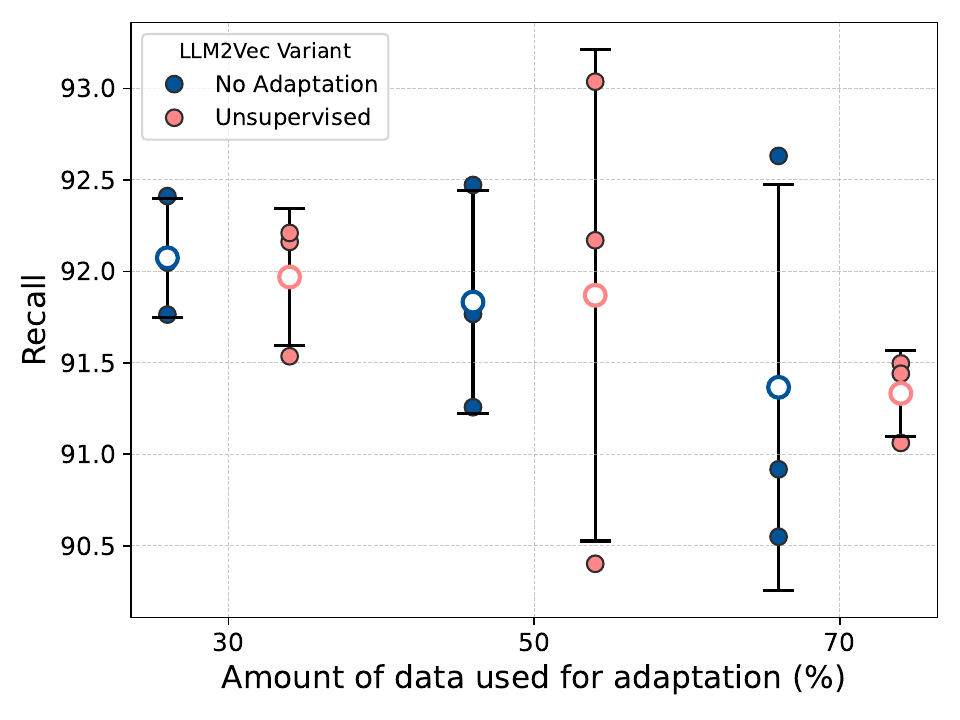}
        \caption{Arabic}
        \label{fig:ft-ar}
    \end{subfigure}
    \begin{subfigure}{0.45\textwidth}
        \centering
        \includegraphics[width=\textwidth]{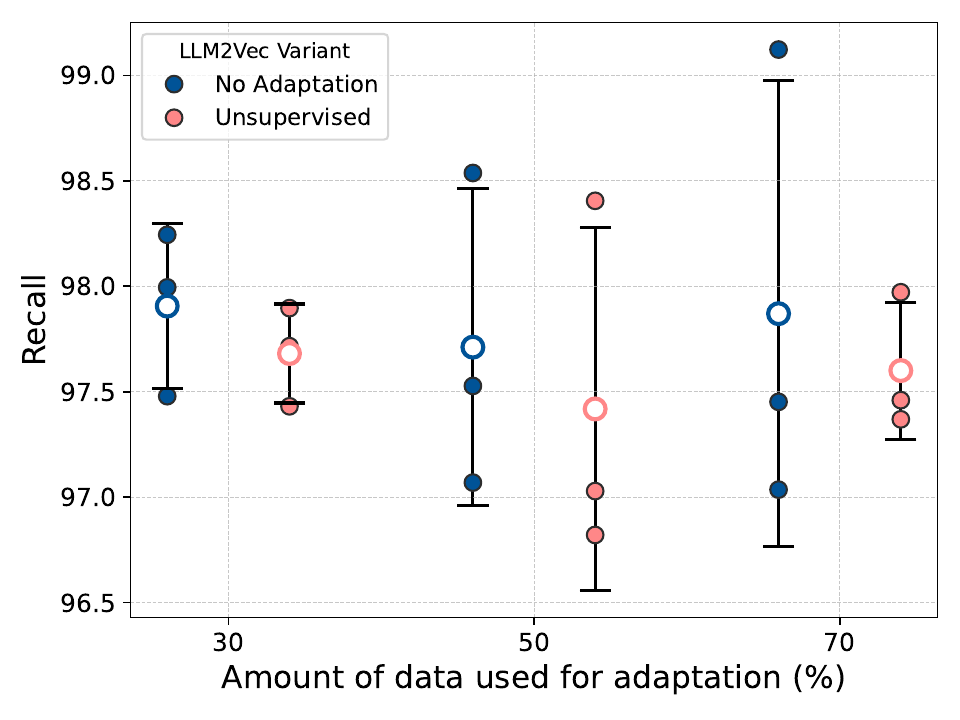}
        \caption{German}
        \label{fig:ft-de}
    \end{subfigure}
     \begin{subfigure}{0.45\textwidth}
        \centering
        \includegraphics[width=\textwidth]{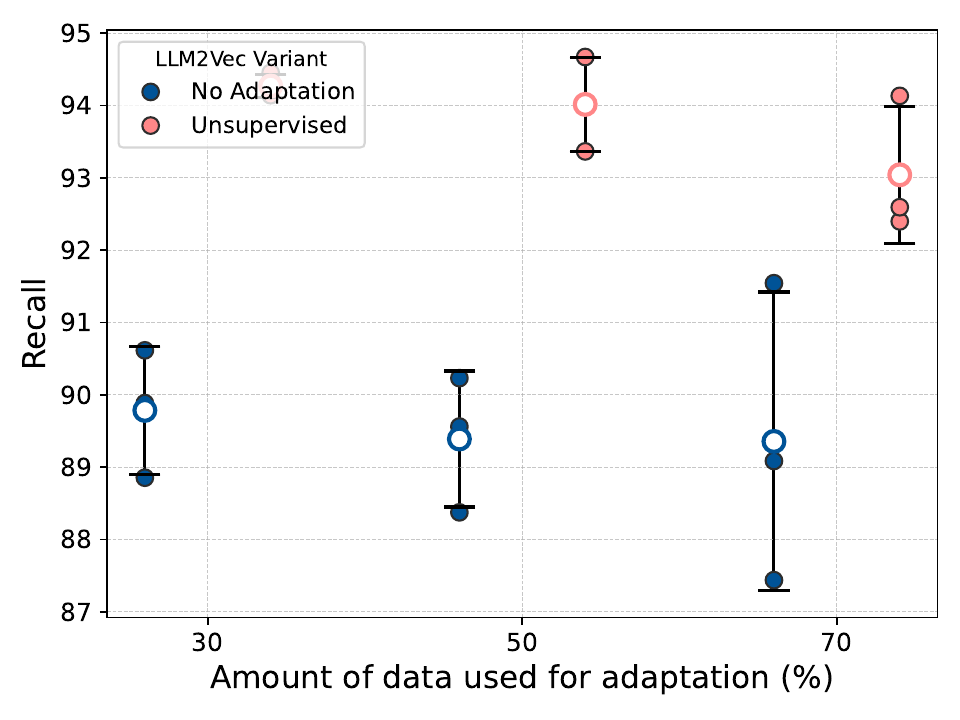}
        \caption{English}
        \label{fig:ft-en}
    \end{subfigure}
    \begin{subfigure}{0.45\textwidth}
        \centering
        \includegraphics[width=\textwidth]{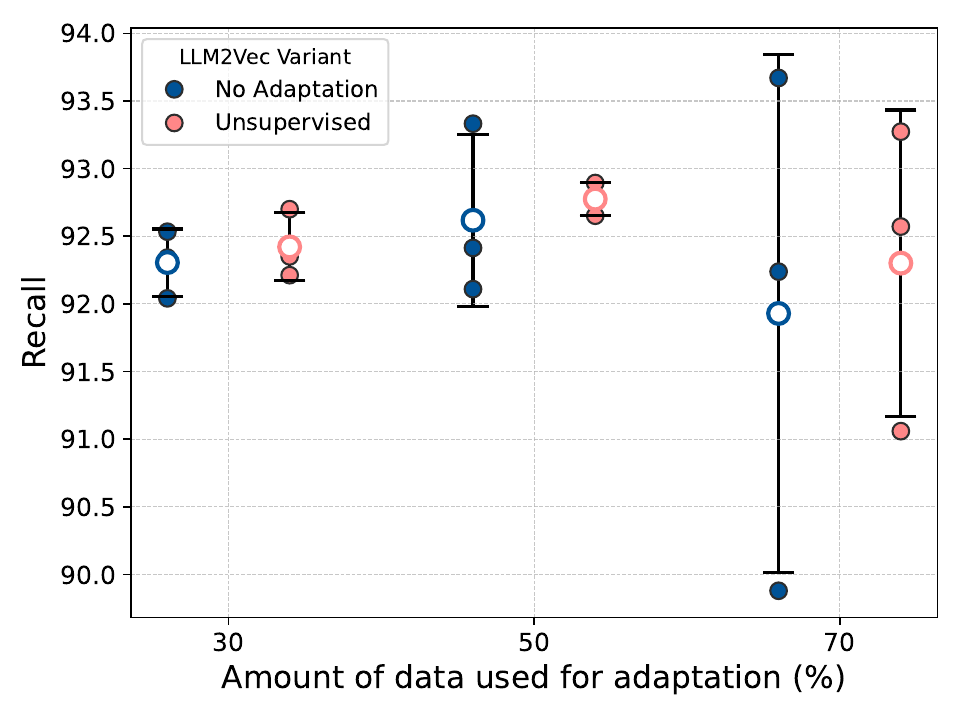}
        \caption{Spanish}
        \label{fig:ft-es}
    \end{subfigure}
    \begin{subfigure}{0.45\textwidth}
        \centering
        \includegraphics[width=\textwidth]{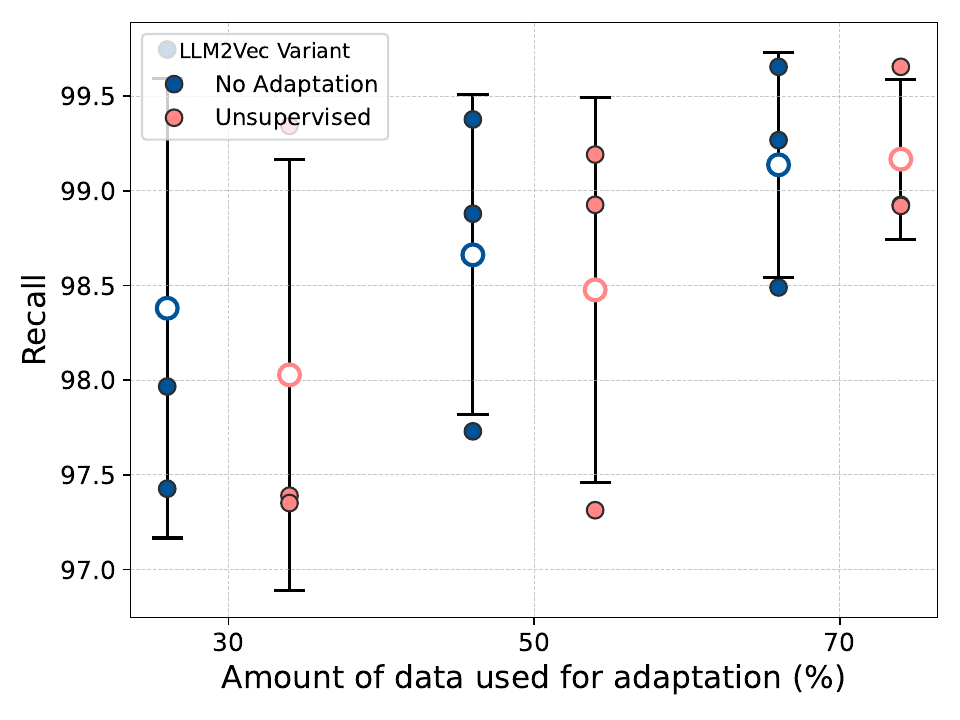}
        \caption{French}
        \label{fig:ft-fr}
    \end{subfigure}
     \begin{subfigure}{0.45\textwidth}
        \centering
        \includegraphics[width=\textwidth]{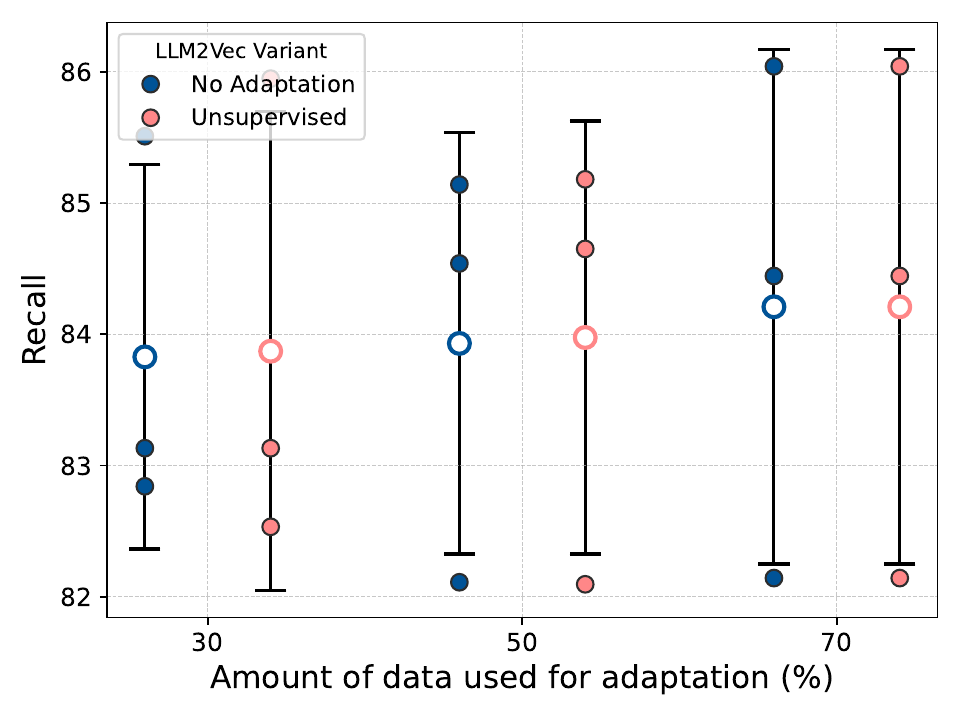}
        \caption{Italian}
        \label{fig:ft-it}
    \end{subfigure}
    \begin{subfigure}{0.325\textwidth}
        \centering
        \includegraphics[width=\textwidth]{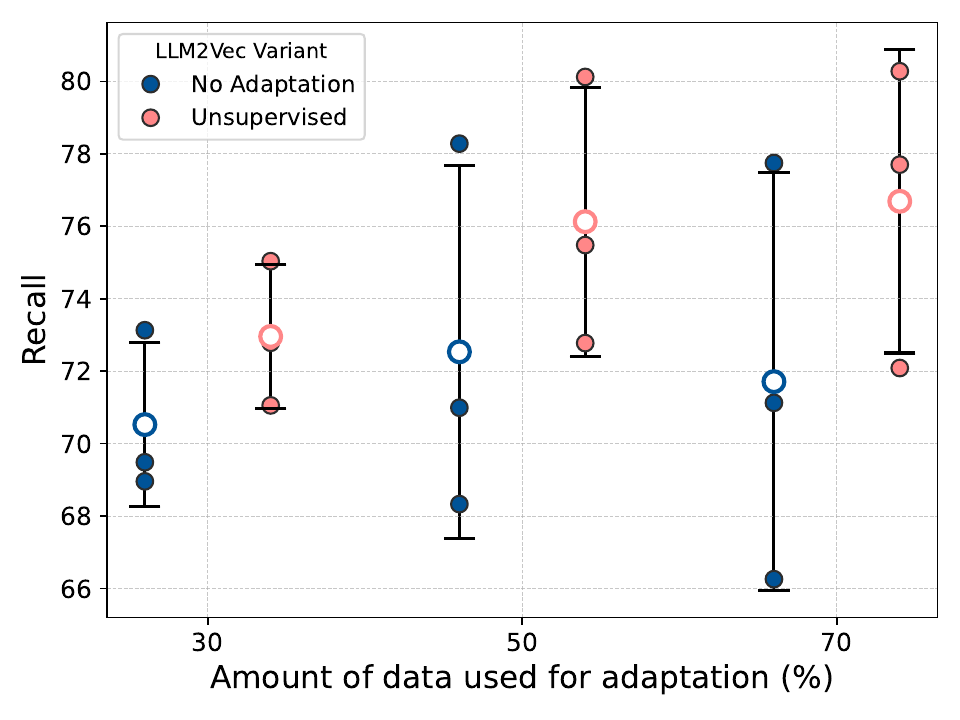}
        \caption{Japanese}
        \label{fig:ft-ja}
    \end{subfigure}
    \begin{subfigure}{0.325\textwidth}
        \centering
        \includegraphics[width=\textwidth]{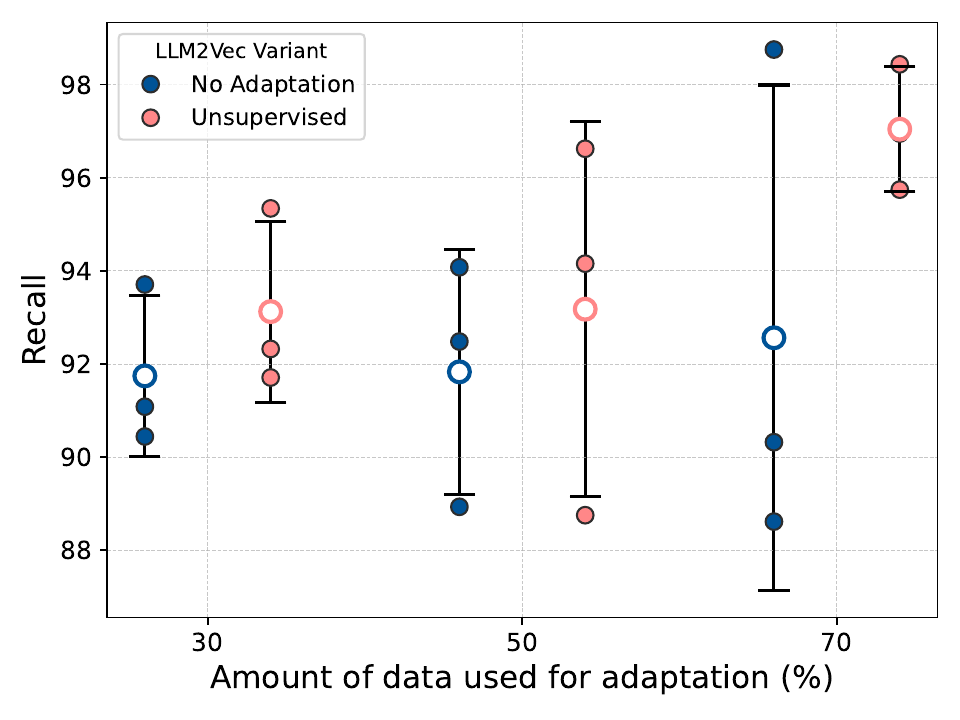}
        \caption{Portuguese}
        \label{fig:ft-pt}
    \end{subfigure}
     \begin{subfigure}{0.325\textwidth}
        \centering
        \includegraphics[width=\textwidth]{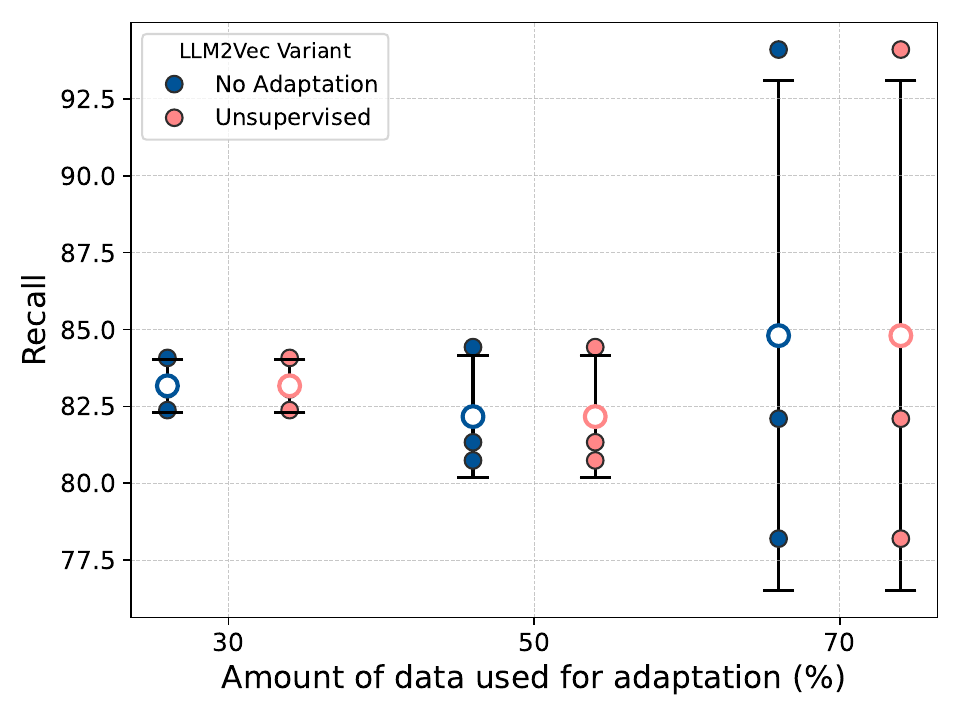}
        \caption{Turkish}
        \label{fig:ft-tr}
    \end{subfigure}
    \caption{Effect of domain adaptation on per-language performance using additional samples from the evaluation set on 3 seeds (solid circles indicate individual runs), including mean (open circle) and standard deviation. Note that the vector space is built using songs from all languages. \textit{No adaptation} corresponds to the original LLM2Vec model, whereas \textit{Unsupervised} performs MNTP-based adaptation. In each scenario, we use Llama 3 8B.
    }
    \label{fig:ft-all-languages}
\end{figure*}

Figure \ref{fig:ft-all-languages} shows results for unsupervised domain adaptation of LLM2Vec using MNTP.
In some languages, such as Italian, French, or Arabic, both models perform similarly. Moreover, we observe a slight difference in Spanish and Portuguese, and a substantial improvement in English and Japanese when using unsupervised MNTP-based domain adaptation.

\end{document}